\documentclass[11pt]{article}
\usepackage{paralist}
\pdfoutput=1

\usepackage[preprint]{acl}

\usepackage{times}
\usepackage{latexsym}
\usepackage[T1]{fontenc}
\usepackage[utf8]{inputenc}
\usepackage{microtype}
\usepackage{inconsolata}

\usepackage{amsmath,amsfonts,amssymb,nicefrac}

\usepackage{graphicx}
\usepackage{subcaption}
\usepackage{placeins}
\usepackage{float}

\usepackage{booktabs}
\usepackage{multirow}
\usepackage{adjustbox}

\usepackage{xcolor}
\definecolor{Gray}{HTML}{D9E8F6}
\usepackage[most]{tcolorbox}
\tcbuselibrary{breakable}
\newtcolorbox{takeaway}{
  breakable,
  colback=Gray!15,
  colframe=Gray!50,
  boxrule=0pt,
  arc=0mm,
  left=1mm,right=1mm,top=1mm,bottom=1mm
}
\newtcolorbox{prompt}{
  breakable,
  colback=cyan!15,
  colframe=cyan!50,
  boxrule=0pt,
  arc=0mm,
  left=1mm,right=1mm,top=1mm,bottom=1mm
}

\usepackage{tikz}
\usepackage{forest}
\usetikzlibrary{shadows.blur}

\usepackage{hyperref}
\hypersetup{
  colorlinks,
  linkcolor=black,
  citecolor=black,
  urlcolor=blue
}
\usepackage{cleveref}

\usepackage{enumitem}

\title{Revisiting Test-Time Scaling: A Survey and a Diversity-Aware Method for Efficient Reasoning}
\author{
  Ho-Lam Chung\textsuperscript{1} \quad
  Teng-Yun Hsiao\textsuperscript{1} \quad
  Hsiao-Ying Huang\textsuperscript{1} \quad
  Chunerh Cho\textsuperscript{1} \\
  Jian-Ren Lin\textsuperscript{1} \quad
  Ziwei Zhang\textsuperscript{1} \quad
  Jhen Hsieh\textsuperscript{1} \quad
  Yun-Nung~Chen\textsuperscript{1} \\[0.5em]
  \textsuperscript{1}National Taiwan University
}

\begin{document}
\maketitle

\begin{abstract}

Test-Time Scaling (TTS) improves the reasoning performance of Large Language Models (LLMs) by allocating additional compute during inference. We conduct a structured survey of TTS methods and categorize them into sampling-based, search-based, and trajectory optimization strategies. We observe that reasoning-optimized models often produce less diverse outputs, which limits TTS effectiveness. To address this, we propose \textbf{ADAPT} (\textbf{A} \textbf{D}iversity \textbf{A}ware \textbf{P}refix fine-\textbf{T}uning), a lightweight method that applies prefix tuning with a diversity focused data strategy. Experiments on mathematical reasoning tasks show that \textbf{ADAPT} reaches 80\% accuracy using eight times less compute than strong baselines. Our findings highlight the essential role of generative diversity in maximizing TTS effectiveness.

\end{abstract}

\section{Introduction}

Large Language Models (LLMs) \cite{openai2023gpt4,chowdhery2022palm,touvron2023llama} have become central to modern NLP applications such as generation, translation, and question answering. Their success largely stems from transformer-based architectures \cite{vaswani2017attention} and large-scale pretraining \cite{kaplan2020scalinglawsneurallanguage, hoffmann2022trainingcomputeoptimallargelanguage}, which endow models with strong fluency and generalization. However, standard autoregressive decoding imposes a fixed inference routine that limits their performance on complex reasoning tasks.

\begin{figure}[ht]
    \centering
    \includegraphics[width=1\linewidth]{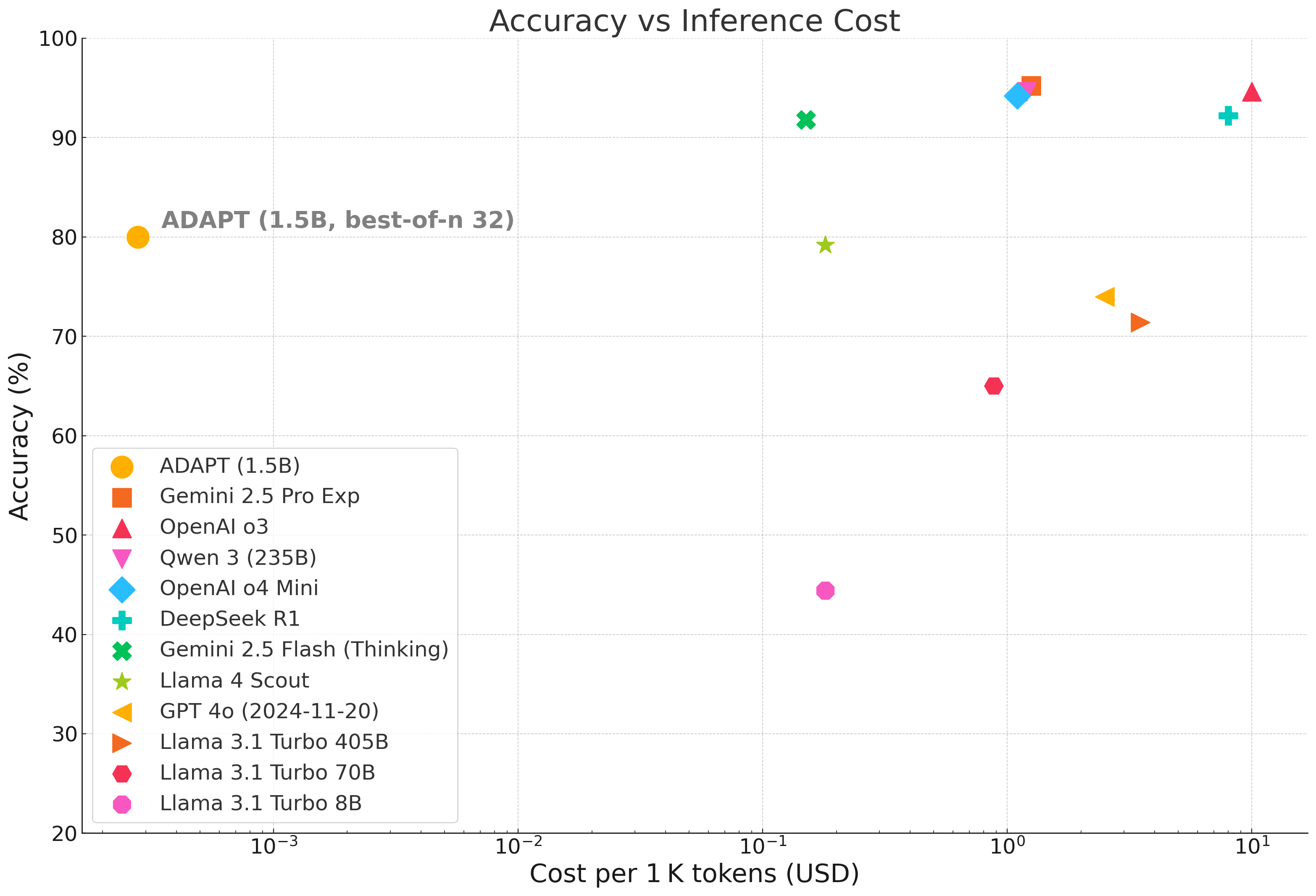}
    \caption{\textbf{Accuracy vs. Inference Cost (log scale).} Each point represents a language model.}
    \label{fig:adapt}
\end{figure}

As model sizes grow, the training cost escalates, yet the marginal gains diminish. To mitigate this, Test-Time Scaling (TTS) has emerged as a promising direction: it enhances model performance by allocating more compute during inference, allowing adaptation to input complexity without retraining \cite{openai2024learning,snell2024scalingllmtesttimecompute,welleck2024decodingmetagenerationinferencetimealgorithms}.

While TTS has shown effectiveness, its performance is often tied to the model's intrinsic capacity for generation diversity—a factor not yet well understood or explicitly optimized. In particular, models optimized for reasoning, such as distilled variants, tend to exhibit reduced output variance, which may dampen the gains from TTS. This raises an open question: Can diversity-aware fine-tuning improve TTS effectiveness for reasoning models?

To address this, we first conduct a strategy-oriented survey of recent TTS methods, categorizing them into three major families: Sampling (\cref{sec:sampling_based}), Search (\cref{sec:search_based}), and Trajectory Optimization (\cref{sec:Trajectory_Optimization}) and identify diversity as a critical enabler of TTS success. Next, we propose a simple yet effective fine-tuning method, ADAPT (A Diversity Aware Prefix fine-Tuning), which enhances early-stage output diversity via prefix-tuned sampling.

We evaluate ADAPT on a compact reasoning model under Best-of-$N$ sampling. As shown in \cref{fig:adapt}, ADAPT achieves 80\% accuracy with eight times fewer samples, outperforming all baseline models in efficiency while retaining strong peak performance.

\paragraph{Contributions.}
This work makes three key contributions:
\begin{compactitem}
\item A unified survey of TTS approaches, covering Sampling, Search, and Trajectory Optimization, with a focus on the role of generation diversity.
\item The design and evaluation of \textbf{ADAPT}, a prefix-tuning method that improves efficiency by increasing diversity at inference.
\item A discussion on future directions, including robustness to prompts, synergy between training and inference, hallucination mitigation, safety, and the use of synthetic data for controlled TTS benchmarking.
\end{compactitem}

\section{Related Work}

Several surveys cover aspects of reasoning, post-training, and test-time scaling.
Reasoning-focused overviews include \citet{pan2025surveyslowthinkingbasedreasoning, xu2025largereasoningmodelssurvey}, while efficiency-oriented surveys include \citet{feng2025efficientreasoningmodelssurvey, wang2025harnessingreasoningeconomysurvey, sui2025stopoverthinkingsurveyefficient}.
Test-time scaling itself is addressed by \citet{zhang2025surveytesttimescalinglarge, li2025a}, and post-training techniques are reviewed in \citet{kumar2025llmposttrainingdeepdive, tie2025surveyposttraininglargelanguage}.

Unlike prior work that focuses primarily on categorization, we go further by connecting our survey insights to a practical hypothesis: that generation diversity moderates TTS effectiveness. We validate this hypothesis through targeted experiments, showing that increasing diversity via prefix tuning leads to more efficient and effective test-time scaling.

\section{Test-Time Scaling Survey}
\label{sec:method}
In this section, we classify TTS methods into three categories, based on the strategies employed.

\subsection{Sampling}
\label{sec:sampling_based}
\begin{figure}[ht]
    \centering
    \includegraphics[width=1\linewidth]{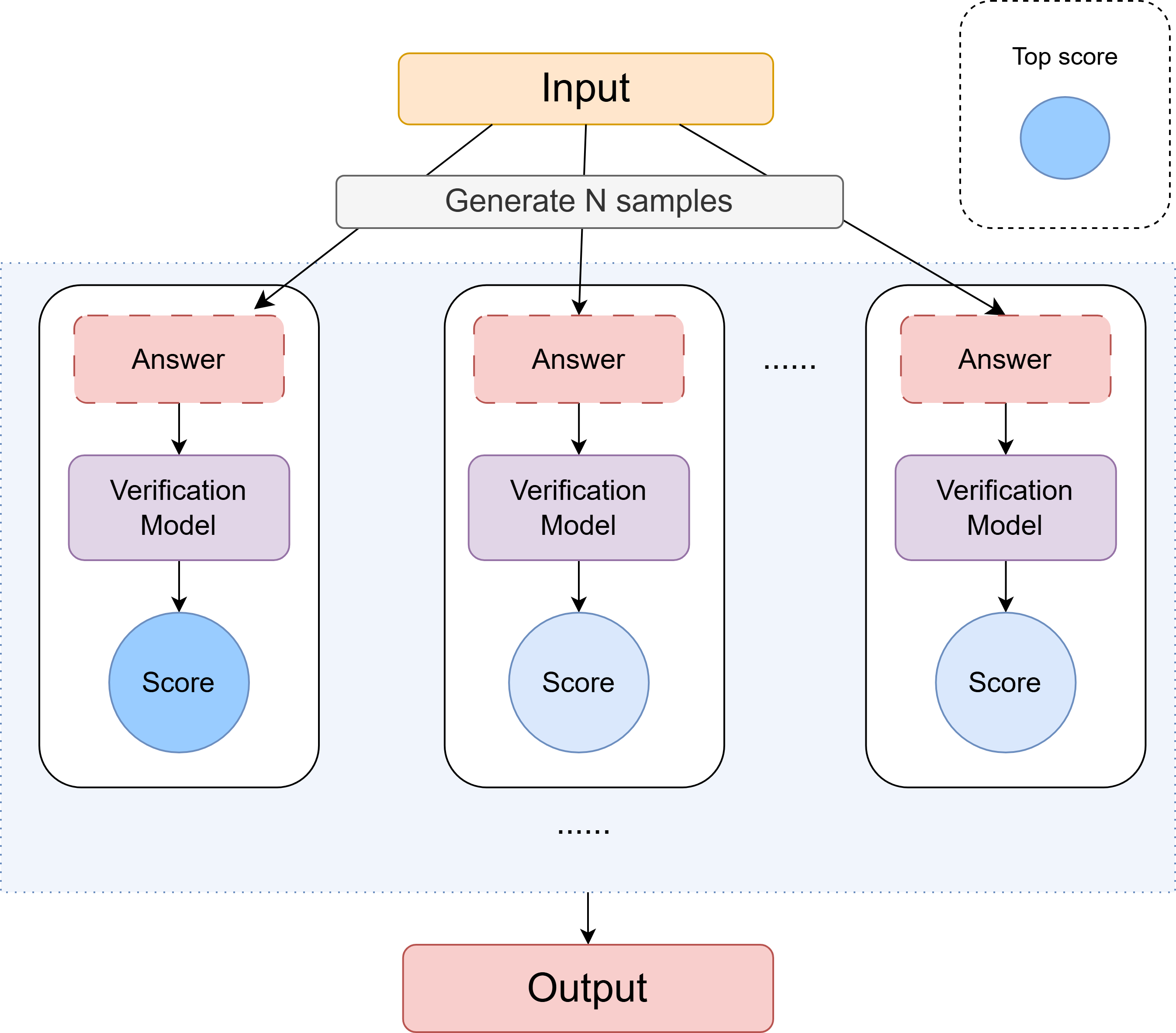}
    \caption{\textbf{Sampling-based method.} The model samples $N$ candidates, a verifier scores each, and the highest‑scoring answer is returned. Some variants repeat this loop for multiple rounds.}
    \label{fig:Sampling-based}
\end{figure}

Sampling methods draw multiple candidates by adjusting decoding parameters such as temperature. For instance, adjusting the temperature redistributes probabilities to favor rare tokens, encouraging more creative responses. After generating the samples, a verifier grades each response and selects the one with the highest score to be the final response. As shown in Figure~\ref{fig:Sampling-based}, the schematic illustrates the basic process of the sampling-based method.

Given the ability to generate diverse responses, numerous previous studies have focused on this and proposed various strategies to improve performance. For example, \citet{tian2025thinktwiceenhancingllm} proposed a method that generates answers in multiple rounds.
In each round, the model uses the previous answer and the original input as a new input.
This approach allows the model to improve its performance without additional training.

In addition, \citet{chow2024inferenceawarefinetuningbestofnsampling} aim to improve LLM performance under Best-of-$N$ sampling through two approaches. BoN-SFT is a supervised fine-tuning method that maximizes the likelihood of the highest-scoring output among sampled candidates, thereby aligning the model with the Best-of-$N$ policy. In contrast, BoN-RL employs policy-gradient reinforcement learning to directly maximize the expected reward of the selected output under Best-of-$N$ inference.

Furthermore, to address efficiency issues faced by sampling-based strategies, \citet{huang2025efficienttesttimescalingselfcalibration} proposed a self-calibrated sampling method.
They used calibrated confidence scores to enhance the efficiency of answer sampling.
The method adjusts sampling through early stopping in Best-of-$N$ and applies confidence-calibrated self-consistency to reduce computation.

\paragraph{The limitations of sampling.}

Although sampling improves performance, \citet{stroebl2024inferencescalingflawslimits} noted that accuracy gains plateau as verifiers become unreliable with more samples. \citet{huang2025bestofnbestthemcoverage} found that Best-of-N can suffer from reward hacking, especially in low-diversity models where sampled responses are redundant. They introduced pessimistic rejection sampling to filter out unreliable outputs, a limitation especially pronounced in distilled models.

\subsection{Search}
\label{sec:search_based}
\begin{figure*}[ht]
    \centering
    \includegraphics[width=0.8\linewidth]{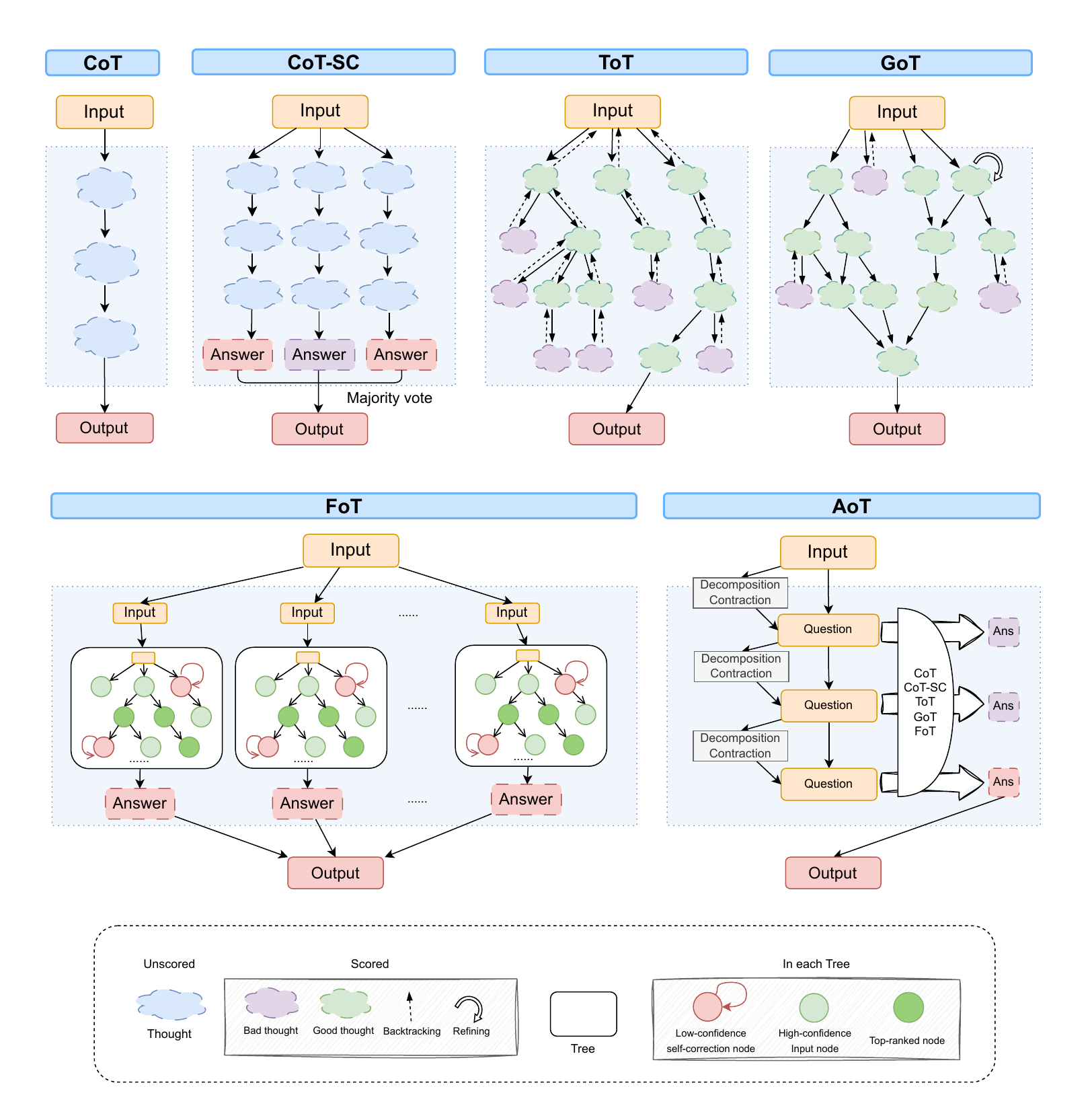}
    \caption{\textbf{Search-Based} The schematic illustrates various search-based methods, including Chain of Thought (CoT), Self-Consistency with Chain of Thought (CoT-SC), Tree of Thoughts (ToT), Graph of Thoughts (GoT), Forest of Thoughts (FoT), and Atom of Thoughts (AoT).}
    \label{fig:search_based}
\end{figure*}
Search-based methods focus on searching for diverse paths.
These paths help improve answer quality.
Some search in the latent space.
This provides an alternative to token-level reasoning.
Others adopt the self-improvement method to enhance their performance.

\subsubsection{Variations of CoT}
\label{sec:variations_of_CoT}
The origins of CoT stem from prompt-based methods \cite{lightman2024lets, ranaldi2025improvingchainofthoughtreasoningquasisymbolic}, such as “Let’s think step by step.” 
Several improved CoT variants are shown in \cref{fig:search_based}.

CoT with Self-Consistency (CoT-SC) \cite{wang2023selfconsistencyimproveschainthought} samples multiple reasoning paths. It then selects the one that best meets a consistency criterion. This improves performance but increases computation.
Auto-CoT \cite{zhang2022automaticchainthoughtprompting} reduces the effect of incorrect demonstrations in the thought process.
Monte Carlo Tree Search (MCTS) \cite{xie2024montecarlotreesearch} updates the model policy through Direct Preference Optimization (DPO) to manage the trade-off between training and inference.

Similarly, the Tree-of-Thoughts (ToT)  \citet{yao2023treethoughtsdeliberateproblem} enables exploration of multiple reasoning paths per step and selects the best action.
Building on multi-path exploration, Forest-of-Thought (FoT) \cite{bi2025forestofthoughtscalingtesttimecompute} runs multiple reasoning trees in parallel instead of a single one.
Each tree independently explores a possible solution path, and dynamically compares, discards, or merges ideas across trees. 
This process is parallel and selective. It improves diversity, adaptability, and resilience. It also avoids local traps found in single-path methods like CoT and ToT.
Graph of Thoughts (GoT) \cite{Besta_2024} represents reasoning as a graph. Nodes are thoughts. 
Edges are logical links. 
This structure supports revisiting, merging, and reusing ideas. 
It enables flexible reasoning and improves coherence and depth in complex tasks.

In addition to structural innovations, Atom of Thoughts (AoT) \cite{teng2025atomthoughtsmarkovllm} breaks down complex tasks into self-contained sub-questions. 
This reduces reliance on historical context and improves generalization.
On the efficiency side, Chain of Draft \cite{xu2025chaindraftthinkingfaster} constrains the length of the thinking trace, improving both speed and output quality. 
Finally, the Mixture-of-Agents architecture \cite{wang2024mixtureofagentsenhanceslargelanguage} leverages multiple LLMs with specialized capabilities to collaborate on complex reasoning tasks, further broadening the design space of CoT-style reasoning.
\paragraph{Theory of CoT.}
While still limited, recent work begins to formalize CoT’s reasoning process beyond empirical improvements.
One approach explores the expressive power of CoT, as studied in
\citet{liu2023transformerslearnshortcutsautomata, merrill2024expressivepowertransformerschain, li2024chainthoughtempowerstransformers, feng2023revealingmysterychainthought}.
Other work focuses on the connection between CoT and in-context learning \citet{huang2025transformerslearnimplementmultistep}, as both improve performance without parameter updates.

\subsubsection{Hidden layer search}
\label{sec:Hidden_layer_based}
Hidden layer search offers new opportunities for efficiency and abstraction.
Sleep-time compute method \cite{lin2025sleeptimecomputeinferencescaling} processes the input context offline.
It generates a latent representation during this stage. 
The model reuses this representation at inference time. 
This method avoids repeating full context processing for each query. 

The Coconut paradigm (Chain of Continuous Thought) \cite{hao2024traininglargelanguagemodels} enables models to reason in a continuous latent space by feeding the last hidden state back into the model instead of decoding it into tokens.
This method allows backtracking and enables reducing token usage significantly.
Latent-Thought Language Models (LTMs) \cite{kong2025scalablelanguagemodelsposterior} guides decoding through latent thought vectors inferred using variational Bayesian techniques. 

Another line of work, CODI \cite{shen2025codicompressingchainofthoughtcontinuous}, introduces a framework that compresses CoT reasoning into a continuous latent space via self-distillation.
Looped transformer \cite{saunshi2025reasoninglatentthoughtspower} presents another hidden-layer based strategy by repeatedly applying a smaller transformer module across iterations. This looping mechanism enables smaller models to match the reasoning capabilities of larger ones on complex reasoning tasks.

\subsubsection{Self-improvement}
\label{sec:Self_improvement_based}
TTS often leverages implicit self-improvement through refinement and correction.
These approaches draw inspiration from human iterative thinking, enabling models to refine their answers dynamically by leveraging internal feedback signals and reward-guided strategies.

Self-Refine \cite{madaan2023self} introduces a simple, training-free framework in which LLMs generate self-criticism and revise their responses accordingly. 
Building on this, STaR \cite{zelikman2022star} bootstraps high-quality rationales that lead to correct answers, using them to fine-tune the model essentially allowing LLMs to “teach themselves” how to reason more effectively.

Expanding further, Training LLMs to Self-Correct via RL \cite{kumar2025training} introduce reward-driven strategies. It proposes a reinforcement-based feedback loop to train LLMs to learn how to revise, achieving significant gains in reasoning tasks. Recursive Introspection (RISE) \cite{qu2024recursive}, on the other hand, formalize multi-turn refinement as a Markov Decision Process (MDP) to iteratively optimize outputs through sequential edits.
After RL training, RISE significantly outperforms Self-Refine and parallel sampling, highlighting the benefits of trajectory-aware self-correction.
Feedback-based Test-Time Training (FTTT) \cite{li2025learning} takes self-improvement a step further by applying gradient-based updates during inference. 
Treating reasoning as a local optimization problem, FTTT adapts model weights based on feedback using a learned optimizer (OPTUNE).
This method achieves robust performance.

\subsection{Trajectory Optimization}
\label{sec:Trajectory_Optimization}
Optimization of TTS in LLMs focuses on controlling and refining the reasoning trajectory. This refers to the sequence of intermediate steps generated during inference.
This manifests as adjusting trajectory length or structure based on task difficulty, balancing accuracy with computational cost. Unlike static decoding, optimized test-time reasoning introduces adaptive mechanisms to improve outcomes under fixed compute budgets.

This optimization addresses two critical challenges. First, \citet{yang2025thinkingoptimalscalingtesttimecompute} demonstrated that longer reasoning trajectories don't uniformly improve results; excessive steps often degrade accuracy through error accumulation. Second, compute allocation at test time can be more effective than scaling model parameters \cite{snell2024scalingllmtesttimecompute}, suggesting trajectory optimization offers a compute-efficient path to enhanced performance.

\subsubsection{RL Paradigms}
We categorize current methods into two paradigms: reinforcement learning (RL) and non-RL approaches. RL enables optimal test-time compute scaling by aligning model outputs with rewards. \citet{setlur2025scalingtesttimecomputeverification,qu2025optimizingtesttimecomputemeta} showed that RL with step-wise feedback or meta-reinforcement fine-tuning (MRT) significantly improves sampling efficiency and test-time performance. 
MRT frames inference as a meta-RL problem, rewarding useful steps toward correct answers to encourage concise reasoning within token limits.

However, recent critiques question RL's necessity and sufficiency. \citet{yue2025doesreinforcementlearningreally} showed that RL mainly reweights the base model’s outputs, boosting performance at low $N$ but lagging at larger $N$ where exploration matters.
Their findings suggest RL narrows the reasoning scope, potentially suppressing valuable but infrequent trajectories. In contrast, distillation-based methods inject new knowledge into student models, expanding reasoning capacity across all $N$ values. \citet{cheng2025stopsummationminformcredit} identified reward hacking in RL with process reward models (PRMs) and propose a framework with min-form credit assignment, achieving high accuracy using only 30\% of the reasoning steps.

\subsubsection{Distillation Paradigms}
Distillation-based methods offer a data-driven alternative to RL for trajectory optimization. Rather than relying on reward shaping, distillation directly transfers structured reasoning from large teacher models to smaller student models. 

Recent work \cite{shridhar2023distillingreasoningcapabilitiessmaller,hsieh2023distillingstepbystepoutperforminglarger,chen2025unveilingkeyfactorsdistilling,ma2025recallreasoningchainofthoughtdistillation} showed that teacher-guided stepwise supervision enables concise, effective reasoning without relying on long inference chains.
The teacher explores diverse reasoning, such as multiple paths and tree-structured CoTs. This is distilled into the student to produce accurate answers with shorter trajectories. 
Distillation improves by adjusting the supervision format \cite{chen2025unveilingkeyfactorsdistilling} or curating datasets that balance trajectory length and informativeness \cite{yin2025wideningdistillationbottleneckreasoning}.
Moreover, distilled models often match or surpass larger models within similar compute budgets \cite{shridhar2023distillingreasoningcapabilitiessmaller,hsieh2023distillingstepbystepoutperforminglarger} and frequently generalize better than RL-trained models, which may overfit to narrow reward signals \cite{yue2025doesreinforcementlearningreally, cheng2025stopsummationminformcredit}. 
Hybrid strategies combining distillation and RL show promise for robust reasoning \cite{liu2025knowledgedistillationtrainingwheels}, though this remains an open research area.

Several methods optimize trajectory length across paradigms. Z1 \cite{yu2025z1efficienttesttimescaling} trains on paired long-short solutions to enable adaptive generation. MRT \cite{qu2025optimizingtesttimecomputemeta} uses episodic rewards to encourage early termination. PURE assigns credit to penalize low-quality steps, reducing verbosity and error-prone reasoning.

\subsection{Challenges and Summary}
Our preceding survey categorizes TTS approaches into three primary dimensions: parallel scaling, sequential scaling, and computational optimization.
A key insight that emerges is the importance of a model’s \emph{inherent generative diversity} in determining TTS effectiveness. By increasing test-time sampling, models can explore a broader range of reasoning trajectories, thereby increasing the likelihood of arriving at a correct solution.

This observation, however, raises a critical question for models specifically optimized for reasoning. Prior work on model specialization (\cref{sec:Trajectory_Optimization}) suggests that targeted training for specific capabilities may narrow a model’s output distribution. Motivated by this, we examine how such specialization affects TTS performance.

We hypothesize that although reasoning-optimized models are skilled at producing accurate responses, this specialization may inadvertently reduce the output diversity needed for effective test-time scaling. 
This potential trade-off between reasoning precision and generative flexibility serves as the foundation for our experimental investigation.

\section{ADAPT: A Diversity Aware Prefix Fine-Tuning Method}
\label{sec:adapt_method}

Motivated by the hypothesis that generative diversity influences TTS performance, we propose a simple yet effective method to encourage diversity and empirically test whether increased diversity leads to improved TTS results. 
Our method, \textbf{ADAPT} (\textbf{A} \textbf{D}iversity-\textbf{A}ware \textbf{P}refix fine-\textbf{T}uning), explicitly promotes generation diversity through an efficient prefix fine-tuning strategy. The key idea is to fine-tune only the initial segments of reasoning trajectories using a carefully curated data mixture, thereby encouraging the model to explore a broader range of initial reasoning paths. This enhanced diversity is expected to enable TTS methods such as best-of-$N$ sampling to identify correct solutions more efficiently, requiring fewer candidates than less diverse models.

\subsection{Dataset Curation}
The training dataset consists primarily of \emph{diverse} responses, supplemented with a smaller subset of outputs generated by the target model, which may exhibit lower generative diversity. This latter subset is included to mitigate potential catastrophic forgetting and to preserve the model's original capabilities.

In our experiments, the dataset includes 90\% responses generated by \texttt{Qwen2.5-Math-1.5B} and 10\% inference outputs from \texttt{DeepSeek-R1-Distill-Qwen-1.5B} (our target model). 
For the Qwen-derived examples, we employ a custom prompt format designed to encourage varied initial reasoning steps (see \cref{sec:appendix:adpatdetails} for details), whereas the DeepSeek-generated samples retain their original chat template.
Since all training targets are produced by existing models, this fine-tuning process can be viewed as a form of targeted knowledge transfer or self-supervised learning.

\subsection{Fine-Tuning Procedure}
\label{ssec:adapt_training}

To improve efficiency and focus on the early stages of reasoning, we truncate all training instances to their first 512 tokens and fine-tune the model using a supervised learning objective. Gradient updates are applied only to the prefix segments, while the remainder of the model remains frozen, preserving most of the pre-trained parameters.

\section{Experiments}
\label{sec:experiments}

In this section, we detail the experimental setup designed to evaluate our proposed method, \textbf{ADAPT}, and to investigate the interplay between solution diversity, trajectory length, and the performance of TTS for reasoning tasks. We first describe the datasets, evaluation protocols, and baseline models. We then present a comparative analysis, focusing on accuracy and computational efficiency.

\subsection{Experimental Setup}
\label{ssec:exp_setup}

\paragraph{Tasks and Datasets.}
Our experiments are conducted on challenging mathematical reasoning benchmarks, akin to MATH-500 \citep{lightman2023lets}.
The goal is to assess the models' ability to generate correct multi-step reasoning paths.

\paragraph{Evaluation Protocol.}
Inspired by the test-time compute setup highlighted by \citet{beeching2024scalingtesttimecompute}, we employ a Best-of-$N$ sampling strategy for all models. For each problem, we generate $N$ candidate solutions, where $N \in \{2, 4, 8, 16, 32, 64, 128, 256\}$. Unless otherwise specified, solutions are generated with a temperature of $0.8$ and a maximum length of $2048$ tokens per problem. The final answer is determined by majority voting over the $N$ candidates, and we report the accuracy based on this aggregated answer (\texttt{acc\_maj}).

\paragraph{Metrics.}
We evaluate model performance using four metrics. \textbf{acc\_maj} denotes the final accuracy obtained via majority voting over $N$ sampled outputs. \textbf{Improvement} measures the absolute increase in \texttt{acc\_maj} relative to the baseline performance at $N=2$. \textbf{Gain per generation} quantifies the average accuracy gain when doubling the sample size (e.g., from $N=2$ to $N=4$). Finally, \textbf{Min $N$ to hit threshold} refers to the smallest sample count $N$ required to reach a target \texttt{acc\_maj}, such as $80\%$.

\subsection{Models}
\label{ssec:models}
Three models are compared in our study: 
\begin{compactitem}
\item \textbf{Qwen-1.5B}: A pre-trained language model tailored for mathematical reasoning, likely to exhibit higher generative diversity. (\texttt{Qwen2.5-Math-1.5B})
\item \textbf{DeepSeek-Qwen-1.5B}: A distilled variant optimized for reasoning tasks~\citep{deepseekai2025deepseekr1incentivizingreasoningcapability}, which may exhibit reduced generative diversity. This model serves as the target for our ADAPT method. (\texttt{DeepSeek-R1-Distill-Qwen-1.5B})
\item \textbf{ADAPT}: A diversity aware prefix fine-tuning approach applied to \textbf{DeepSeek-Qwen-1.5B}, aimed at enhancing its generative diversity while preserving reasoning accuracy.
\end{compactitem}

\subsection{Diversity Impact on TTS Performance}
\label{ssec:tts_challenges}

We begin by examining how the effectiveness of best-of-$N$ sampling in TTS depends on both solution diversity and trajectory length.
We compare a pre-trained model (\texttt{Qwen2.5-Math-1.5B}) with its distilled counterpart optimized for reasoning (\texttt{DeepSeek Qwen-1.5B}).

\begin{figure}[t!]
    \centering
    \includegraphics[width=\linewidth]{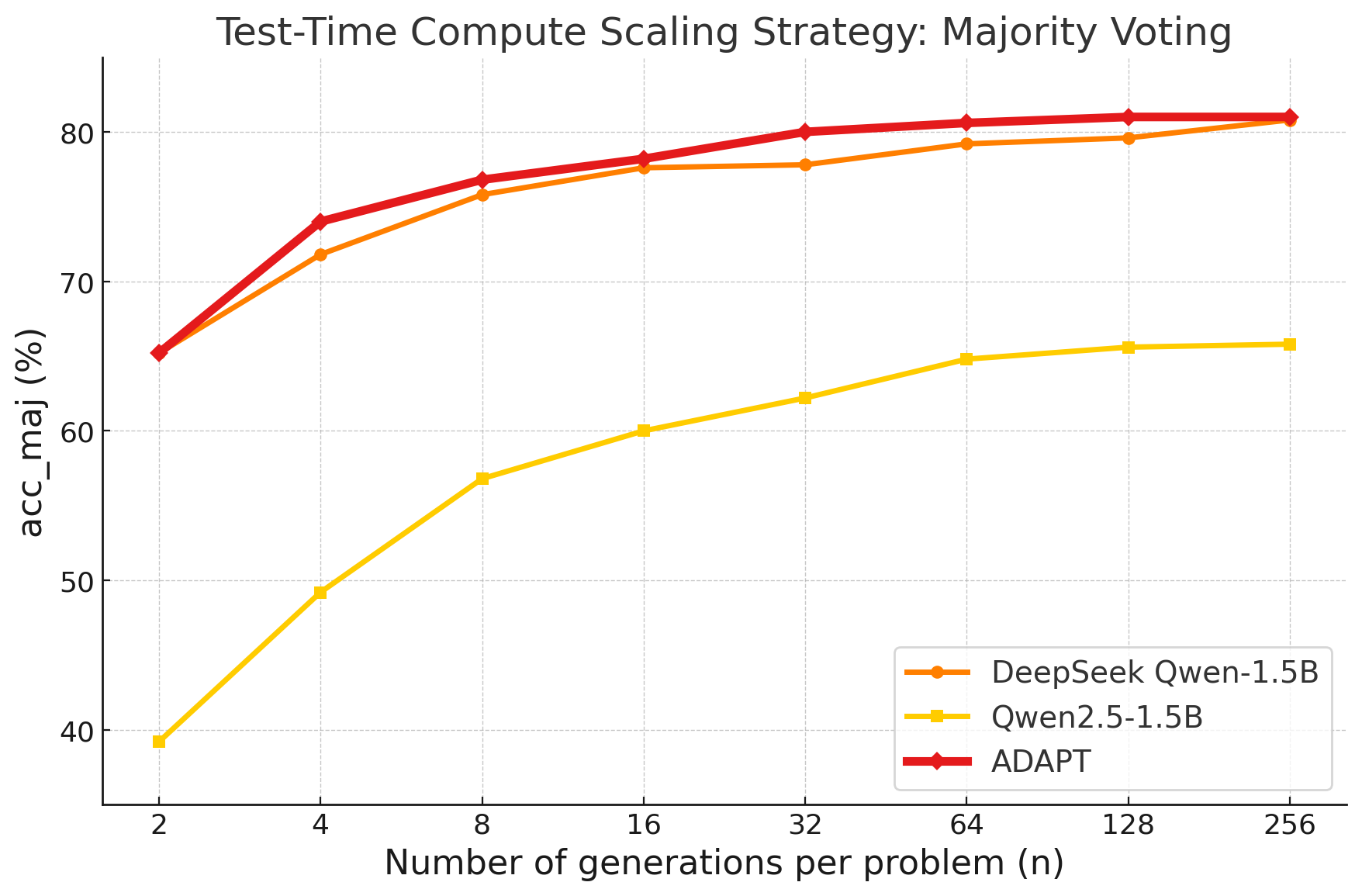}
    \caption{\textbf{TTS performance comparison of pre-trained vs. distilled model.}}
    \label{fig:initial_comparison_placeholder}
\end{figure}

As shown in \autoref{fig:initial_comparison_placeholder}, \texttt{DeepSeek Qwen-1.5B} achieves a higher baseline accuracy of 65.2\% at $N=2$, reaching 80.8\% at $N=256$ (+15.6\%).
In contrast, \texttt{Qwen2.5-Math-1.5B} starts lower at 39.2\% but exhibits a steeper improvement, reaching 65.8\% (+26.6\%).
These findings suggest that while distillation enhances baseline reasoning performance, it may also suppress generative diversity, limiting the benefits of increased sampling.
This observation well motivated our hypothesis: explicitly promoting diversity during fine-tuning can enhance both the efficiency and effectiveness of TTS for reasoning-specialized models.

\subsection{Results}
\label{ssec:main_results}

\autoref{tab:performance_comparison} summarizes the results of \textbf{ADAPT} compared to both baselines. Notably, \textbf{ADAPT} reaches $80\%$ accuracy with only $32$ samples, an $8\times$ improvement over \texttt{DeepSeek Qwen-1.5B}. At $N=16$, it already surpasses the distilled baseline at $N=256$.

\begin{table*}[t!]
\centering
\begin{tabular}{lcccc}
\toprule
\bf Model   & \bf Acc. Maj@2 & \bf Acc. Maj@256 & \bf Acc. Maj@16 & \bf Min $N$ for 80\% ($\downarrow$) \\
\midrule
Qwen2.5-1.5B            & 39.2\%   & 65.8\%  & 60.0\%    & $\infty$           \\
DeepSeek Qwen-1.5B      & 65.2\%  & 80.8\%  & 77.6\%  & $256$              \\
ADAPT (Ours)   & \bf 65.2\%   & \bf 81.0\%    & \bf 78.2\%    & \bf ~~32         \\
\bottomrule
\end{tabular}
%\end{adjustbox}
\caption{\textbf{Best-of-$N$ majority voting results.} ``Min $N$'' indicates the smallest $N$ required to reach $80\%$ accuracy.}
\label{tab:performance_comparison}
\end{table*}

As shown in \autoref{tab:performance_comparison}, both \textbf{ADAPT} and \texttt{DeepSeek Qwen-1.5B} achieve strong initial performance ($65.2\%$ at $N=2$), clearly outperforming the pre-trained \texttt{Qwen2.5-1.5B} ($39.2\%$). \textbf{ADAPT} attains the highest peak accuracy ($81.0\%$ at $N=256$), marginally surpassing \texttt{DeepSeek Qwen-1.5B} ($80.8\%$). While \texttt{Qwen2.5-1.5B} shows the largest relative improvement (+26.6 pp), its ceiling remains substantially lower.

Beyond peak accuracy, \textbf{ADAPT} offers significant advantages in convergence speed and sampling efficiency. It reaches $80\%$ accuracy with just $N=32$ samples—an 8× improvement over the $N=256$ needed by \texttt{DeepSeek Qwen-1.5B}. With only $N=16$ samples, \textbf{ADAPT} already achieves $78.2\%$, making it suitable for low-budget inference. In contrast, \texttt{Qwen2.5-1.5B} fails to reach the $80\%$ threshold even at $N=256$.

\begin{figure}[t!]
    \centering
    \includegraphics[width=\linewidth]{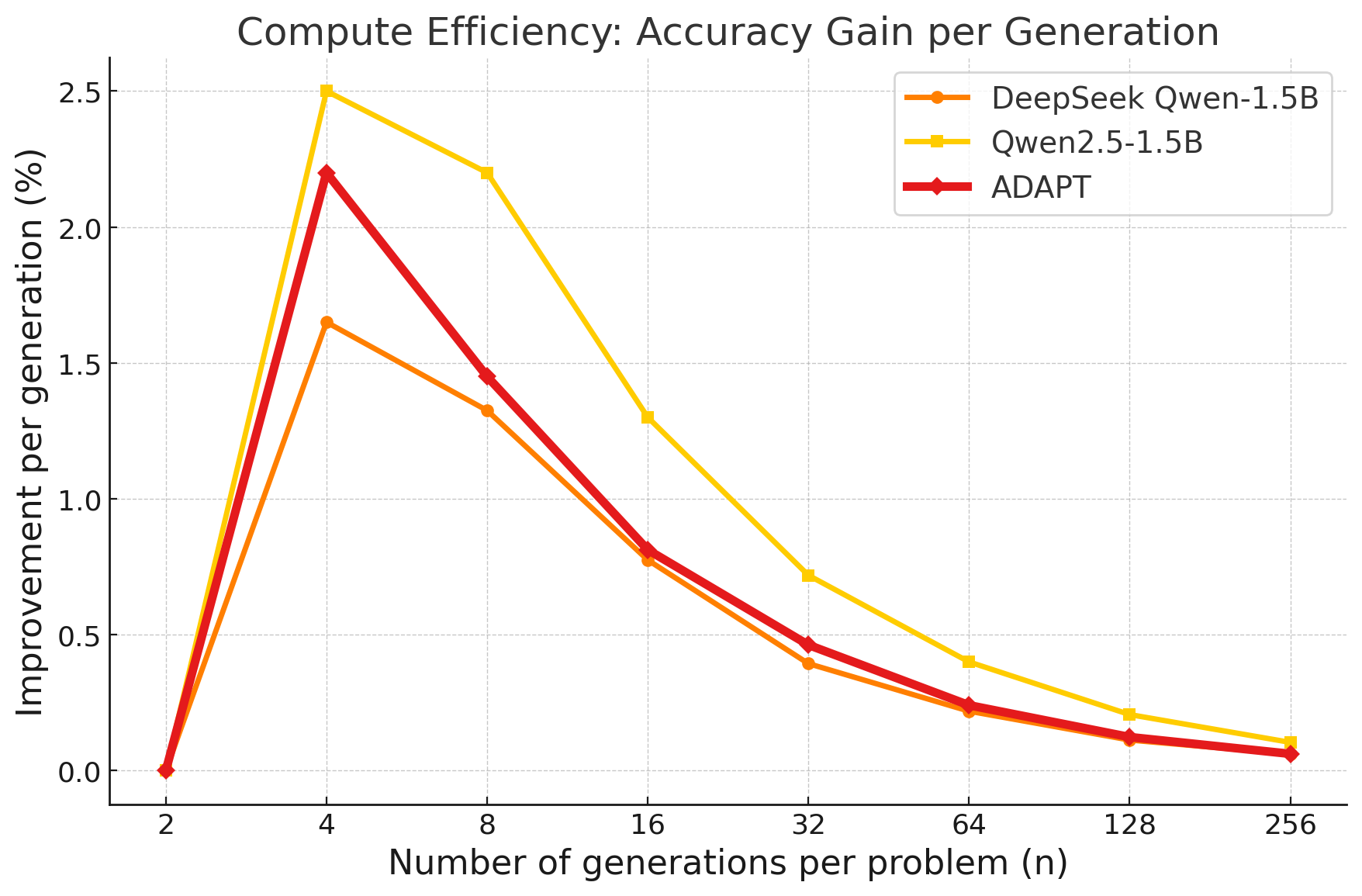} % Make sure this path is correct
    \caption{\textbf{Marginal gain per generation for across models}}
    \label{fig:gain_per_generation_placeholder}
\end{figure}

\autoref{fig:gain_per_generation_placeholder} illustrates the diminishing returns of larger $N$ for all models. For \textbf{ADAPT}, most gains occur early ($N=2$ to $N=32$), suggesting that it captures the majority of TTS benefits with a relatively small number of generations.

\subsection{Discussion}
\label{ssec:discussion}

Our results validate the central hypothesis: enhancing output diversity improves both the accuracy and efficiency of reasoning models under Best-of-$N$ sampling. While \texttt{DeepSeek Qwen-1.5B} offers a strong baseline, its improvements require large $N$, consistent with prior observations that distillation, though beneficial for core reasoning, may suppress generative diversity.

\textbf{ADAPT} addresses this limitation via prefix fine-tuning focused on early-stage reasoning. By updating only a small number of prefix parameters and training on a hybrid dataset—composed of model-generated responses and a subset of base model outputs—ADAPT promotes diverse initial reasoning paths while mitigating catastrophic forgetting. The prompt template further steers variation in the early solution space (see \cref{sec:appendix:adpatdetails}).

This strategy yields strong gains even under limited sampling budgets: \textbf{ADAPT} exceeds $80\%$ accuracy with only $N=32$ samples and achieves $78.2\%$ at $N=16$. These results demonstrate that diversity-aware prefix tuning enables more efficient exploitation of TTS compared to standard fine-tuned models.

In summary, ADAPT delivers faster convergence, stronger low-sample performance, and better overall accuracy under Best-of-$N$ majority voting. Its improvements stem from explicitly optimizing for sample-efficient diversity, making it a compelling approach for resource-constrained inference.

\section{Future Directions}
In this section, we outline several future directions in the TTS regime.
As a new research area, TTS still faces many unexplored challenges and developed applications. A detailed discussion of TTS applications is provided in \cref{sec:appendix}.
Here, we focus on five key categories: robustness, training, safety, hallucination, and synthetic datasets.

\paragraph{Robustness.}
TTS might fail under certain conditions. For instance, prompt format significantly affect LLM performance on the same task. \citet{hochlehnert2025soberlookprogresslanguage} showed that reasoning accuracy is highly sensitive to prompt design. 
Therefore, investigating how the content and structure of prompts affect TTS is a promising direction for future research.

\paragraph{Training.}
\citet{dang2025weightensemblingimprovesreasoning} identified that the training of the reasoning models leads to suboptimal performance of TTS.
Furthermore, \citet{chen2025rethinkingfinetuningscalingtesttime} suggested that modifications to the pre-training / fine-tuning stages are necessary to optimize TTS performance. 

Specifically, they observe that training with cross-entropy loss leads to a decrease in pass@N accuracy. 
It may result from cross-entropy loss leading to model overconfidence.
The finding underscores the need to reconsider the pre-training and fine-tuning strategies in order to achieve better performance and efficiency.
For instance, it is important to explore the impact of training strategies, such as backpropagation, on TTS performance. 
\paragraph{Safety.}
\citet{chenreasoning} found that the thinking process of LLMs may not be entirely transparent. 
They prompt the model with a question and the corresponding answer, but observe that LLMs rarely acknowledge receiving the hint. 
Additionally, they note that CoT monitoring may not be sufficiently reliable to capture the true cognitive process of LLMs.
Future work should further explore how LLMs perform reasoning to derive answers, in order to establish better control over their behavior.
\paragraph{Hallucination.}
Despite TTS success on some math and coding tests, \citet{openai2024o3o4mini} showed that GPT o3 and o4-mini suffer from the hallucination problem. 
Future work should aim to understand the relationship between RL and SFT methods used to enable LLMs to scale test-time computation, as well as their impact on hallucination.

\paragraph{Synthetic dataset.}
Synthetic datasets \cite{goldie2025syntheticdatageneration,wang2025graphbasedsyntheticdatapipeline,lei2024s3evalsyntheticscalablesystematic} provide precise control over task structure and difficulty. It enables the isolation of specific factors relevant to test-time computation. 
This reduces confounding effects present in natural data and helps reveal how scaling the inference budget impacts reasoning depth, compositionality, and context integration.

\section{Conclusion}
In this paper, we present a comprehensive survey of test-time scaling (TTS), categorizing recent approaches into three main strategies: sampling, search, and trajectory optimization. Through this analysis, we hypothesize that \emph{generative diversity} is a key factor influencing TTS performance. 
Our experiments support this hypothesis, showing that while distillation enhances baseline reasoning accuracy, it can also reduce output diversity, thereby limiting the effectiveness of sampling-based TTS methods. 
To address this limitation, we introduce \textbf{ADAPT}, a diversity aware prefix fine-tuning method designed to enhance output diversity and improve both the efficiency and performance of reasoning-optimized models under TTS.

\section*{Limitations}
\label{sec:limitations}

While \textbf{ADAPT} demonstrates strong performance under Best-of-$N$ sampling, several limitations remain. First, all experiments are conducted on a single reasoning domain—mathematical problem solving. It remains unclear whether similar diversity-induced gains would generalize to broader tasks such as commonsense or multi-hop QA. Second, our evaluations focus on a relatively small model (\texttt{1.5B} parameters); scaling effects and interactions with larger architectures are left for future work.

Third, although ADAPT improves sample efficiency, it does not directly optimize diversity metrics (e.g., self-BLEU, pairwise entropy), and its diversity-enhancing effect is inferred only through indirect accuracy gains. Explicit diversity measurements could provide more rigorous support for the core hypothesis. Finally, we fix the prefix length and data mixture ratio throughout; exploring how these hyperparameters impact diversity and performance may yield further improvements.

\bibliography{custom}

\newpage
\appendix
\section{Appendix}
\label{sec:appendix}
%
%This is an appendix.
\subsection{Methodology Structure}
\label{sec:survey_struc}
% ---------- color & style definitions（如已在前文定義，可省略） ----------
\definecolor{tnodeColor}{HTML}{FFF0C5}
\definecolor{edgeColor}{HTML}{000000}
\definecolor{textColor}{HTML}{000000}

% 若尚未載入：
\usetikzlibrary{shadows.blur}
% ---------- tree diagram ----------
% \begin{figure*}[p]
\begin{center}
% \begin{center}
\centering
\resizebox{1\textwidth}{!}{
\begin{forest}
  %----- global node style -------------------------------------------------
  for tree={
    draw=edgeColor,
    thick,
    font=\sffamily,
    fill=tnodeColor,
    rectangle,
    rounded corners=4pt,
    text=textColor,
    blur shadow={shadow scale=0.95,
                 shadow xshift=.5ex,
                 shadow yshift=-.5ex,
                 shadow opacity=0.25},
    edge={-,draw=edgeColor,line width=1.5pt},
    grow=0,
    child anchor=west,
    parent anchor=east,
    anchor=west,
    align=center,
    l sep+=0.3cm,
    s sep+=0.5cm,
  },
  root/.style={fill=Gray!15, font=\bfseries\huge,
               rounded corners=6pt, text width=3cm,
               /tikz/align=center, inner sep=6pt},
  lvl2/.style={fill=Gray!15, font=\bfseries\large,
               text width=3cm, /tikz/align=center,
               inner sep=6pt},
  lvl2/.style={fill=tnodeColor, font=\sffamily\small,
               text width=4cm, /tikz/align=center,
               inner sep=4pt},
  lvl3/.style={fill=tnodeColor, font=\sffamily\small,
               text width=13cm, /tikz/align=center,
               inner sep=4pt},
  lvl5/.style={fill=tnodeColor, font=\sffamily\small,
               text width=16cm, /tikz/align=center,
               inner sep=4pt},
  %-----------------------------------------------------------------------
  [Methods, root
    % 依顯示順序「由上到下」→ 程式碼需倒排
      [Trajectory Optimization,     lvl2
        [Optimal scaled length of CoT: \citet{yang2025thinkingoptimalscalingtesttimecompute}; \\Test time scaling: \citet{snell2024scalingllmtesttimecompute}; \\Verifier-based is better: \citet{setlur2025scalingtesttimecomputeverification}; \\Meta Reinforcement: \citet{qu2025optimizingtesttimecomputemeta}; RL limit: \citet{yue2025doesreinforcementlearningreally}; \\PURE framework: \citet{cheng2025stopsummationminformcredit}; \\Optimize trajectory length: \citet{yu2025z1efficienttesttimescaling}; \\Segment reasoning: \citet{qu2025optimizingtesttimecomputemeta}; \\, lvl3]
      ]
      [Search,           lvl2
        [Variations of CoT \\
            CoT: \citet{lightman2024lets, ranaldi2025improvingchainofthoughtreasoningquasisymbolic}; \\ 
            CoT with Self-Consistency (CoT-SC): \citet{wang2023selfconsistencyimproveschainthought}; \\ 
            Auto-CoT: \citet{zhang2022automaticchainthoughtprompting};\\
            Monte Carlo Tree Search (MCTS): \citet{xie2024montecarlotreesearch}; \\
            Tree-of-Thoughts (ToT): \citet{yao2023treethoughtsdeliberateproblem} ; Forest-of-Thought (FoT): \citet{bi2025forestofthoughtscalingtesttimecompute}; 
            \\Atom-of-Thoughts (AoT): \citet{teng2025atomthoughtsmarkovllm};\\
            Chain of Draft:  \citet{xu2025chaindraftthinkingfaster};  \\Mixture of Agents: \citet{wang2024mixtureofagentsenhanceslargelanguage};\\, lvl3
        ]
        [Theory of CoT \\ 
            Expressive power of CoT: \citet{liu2023transformerslearnshortcutsautomata, merrill2024expressivepowertransformerschain, li2024chainthoughtempowerstransformers};\\
            Connection between CoT and in-context learning:  \citet{huang2025transformerslearnimplementmultistep};,lvl3]
        [Hidden layer search\\
            Sleep-time compute method: \citet{lin2025sleeptimecomputeinferencescaling};\\
            Coconut paradigm (Chain of Continuous Thought): \citet{hao2024traininglargelanguagemodels};\\
            Latent-Thought Language Models (LTMs):  \citet{kong2025scalablelanguagemodelsposterior};\\
            CODI: \citet{shen2025codicompressingchainofthoughtcontinuous};\\
            Looped transformer: \citet{saunshi2025reasoninglatentthoughtspower};\\
        , lvl3]
        [Self-improvement \\
           Self-Refine: \citet{madaan2023self};\\
            Coconut paradigm (Chain of Continuous Thought): \citet{hao2024traininglargelanguagemodels};\\
            Latent-Thought Language Models (LTMs):  \citet{kong2025scalablelanguagemodelsposterior};\\
            CODI: \citet{shen2025codicompressingchainofthoughtcontinuous};\\
            Looped transformer: \citet{saunshi2025reasoninglatentthoughtspower};\\
        , lvl3]
      ]
      [Sampling,  lvl2
        [Think multiple rounds: \citet{tian2025thinktwiceenhancingllm};\\ 
        Inference-Aware: \citet{chow2024inferenceawarefinetuningbestofnsampling}\\
        Self-Calibration: \citet{huang2025efficienttesttimescalingselfcalibration};\\
        , lvl3]
      ]
  ]
\end{forest}
}

% \caption{Survey strucuture.}
% \captionsetup{type=figure, justification=centering}
% \captionof{figure}{Survey structure.}
% \label{fig:survey-structure}
% \end{figure*}
\end{center}
\begin{minipage}{\textwidth}
  \centering
  
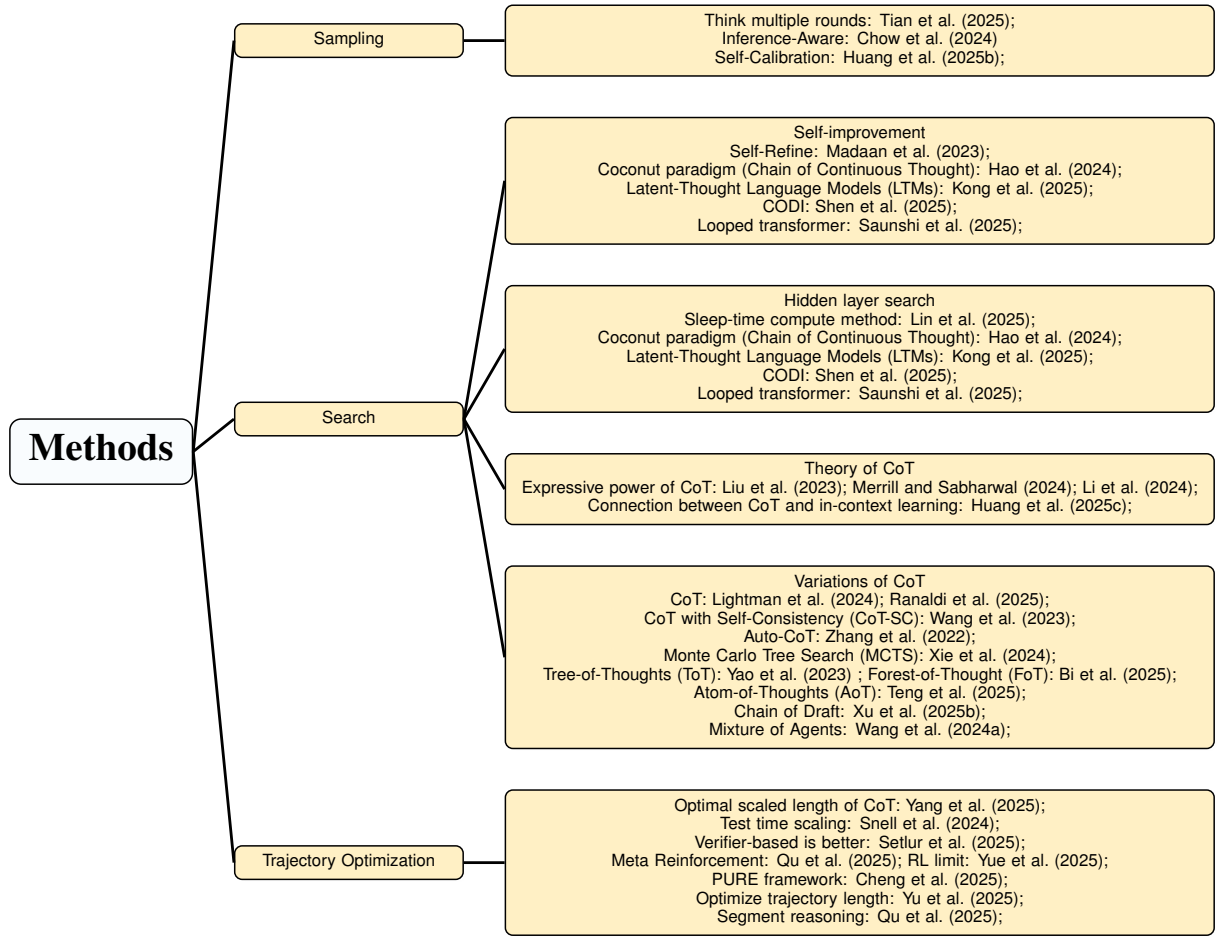
\captionof{figure}{Methodology classification structure.}
  \label{fig:survey-structure}
\end{minipage}

\clearpage
\subsection{Related Work}
There are numerous surveys focusing on various aspects of reasoning, post-training, and TTS. 
\citet{pan2025surveyslowthinkingbasedreasoning, xu2025largereasoningmodelssurvey} survey reasoning models. Efficiency-focused surveys include \citet{feng2025efficientreasoningmodelssurvey}, \citet{wang2025harnessingreasoningeconomysurvey}, and \citet{sui2025stopoverthinkingsurveyefficient}. 
TTS is covered by \citet{zhang2025surveytesttimescalinglarge} and \citet{li2025a}. Post-training techniques are reviewed in \citet{kumar2025llmposttrainingdeepdive} and \citet{tie2025surveyposttraininglargelanguage}.

Unlike previous work that systematically categorizes TTS methods, our study not only investigates TTS classification but, more importantly, proposes an experimental investigation into the trade-offs between reasoning proficiency and output diversity, inspired by our survey.
\subsection{Applications in Real-World Domains}
\paragraph{Robotics and Autonomous Systems} In robotics, TTS facilitates improved decision-making and adaptability in dynamic environments \cite{zawalski2025roboticcontrolembodiedchainofthought}. 
World Foundation Models (WFMs) \cite{nvidia2025cosmosworldfoundationmodel} simulate physical environments. They support tasks like autonomous driving and robotics. TTS improves their prediction. It also increases adaptability during inference.
% For instance, the SWIFT framework integrates TTS to improve simulation fidelity and agent training in robotics applications .

\paragraph{Software Engineering and Autonomous Agents} 
In software engineering, TTS enhances reasoning in autonomous agents handling complex development tasks. 
The SWE-Reasoner framework \cite{ma2025thinkinglongerlargerenhancing} introduce a unified approach that combines internal and external test-time compute strategies to dynamically allocate computational resources during inference. Internally, it leverages development-contextualized long Chain-of-Thought trajectories to to guide smaller models for multi-step reasoning tasks. Externally, it incorporates reward-guided search and execution-based verification to focus inference-time compute on critical development phases such as fault localization and patch generation. 
% This allows a 32B open-source model to achieve a 46\% issue resolution rate on SWE-bench Verified, surpassing much larger models like DeepSeek R1 (671B). 
Complementing this, multi-agent collaborative systems such as M1 + CEO (\cite{jin2025headsbetteronetesttime}) use a central coordination agent to dynamically manage reasoning depth, agent iteration, and communication strategies at test time. 

\paragraph{Video Processing and Streaming Analytics} TTS has shown promise in video processing \citet{dalal2025oneminutevideogenerationtesttime}, particularly in scenarios requiring real-time analysis and adaptation. The Test-Time Training (TTT) approach extends TTS to streaming video data. It helps models adapt to temporal changes. It also improves performance on tasks like instance segmentation and panoptic segmentation.
TTT outperforms traditional fixed-model baselines by continuously updating the model with incoming video frames, thus enhancing accuracy in dynamic visual environments .

\paragraph{Medical Diagnostics and Clinical Decision Support} 
TTS enhances the diagnostic capabilities by allowing more extensive reasoning during inference. 
\cite{huang2025m1unleashpotentialtesttime} demonstrate that increasing the reasoning token budget at test time significantly improves performance on medical question-answering tasks.  
However, the study also notes an optimal reasoning token budget, beyond which performance may degrade due to overthinking.

Additionally, TTS contributes to uncertainty estimation in medical image segmentation \citet{ma2024testtimegenerativeaugmentationmedical}. By applying test-time augmentation techniques, models can better assess aleatoric uncertainty, leading to more reliable segmentation outputs and reducing overconfident incorrect predictions.

\paragraph{Legal Document Analysis and Compliance} Legal document analysis involves processing complex and lengthy texts, where TTS enhances the comprehension and reasoning abilities of AI models \citet{kant2024equitableaccessjusticelogical}. By allocating more computational resources during inference, models can better understand intricate legal language, identify relevant precedents, and ensure compliance with regulations. This capability is particularly valuable in tasks such as contract analysis, legal research, and compliance monitoring.

\begin{takeaway}
\paragraph{Takeaways.}
TTS enhances real-world applications by enabling adaptive, context-aware inference. In robotics, it improves decision-making in dynamic environments. In software engineering, it supports multi-step reasoning and verification in development tasks. For video, it adapts to streaming input for more accurate segmentation. In medicine, it boosts clinical QA performance and improves uncertainty estimation. In legal analysis, it helps models process complex language and ensure compliance.
\end{takeaway}
\subsection{Dataset}
As shown in Table \cref{scc:dataset}, we have collected and summarized several common datasets used in previous work for future experimental implementation.

\begin{table*}[htbp]
\centering
\begin{adjustbox}{max width=\textwidth}
    \begin{tabular}{lcclcccc}
    \toprule
    \bf Dataset & \bf Size & \bf Domain & \bf Description & Ref \\
    \midrule
    DeepScaleR-Preview-Dataset & 40K+ & Math & Problem-answer pairs & \citet{deepscaler2025}\\ 
    MATH & 12.5K & Math & Competition mathematics problems & \citet{hendrycksmath2021}\\
    GSM8K & 8.5K & Math & Grade school math problems & \citet{cobbe2021trainingverifierssolvemath}\\
    GSM-hard & 1.32K & Math & Question-answer pairs & \citet{gao2023palprogramaidedlanguagemodels}\\ 
    GSM-Symbolic & 5K & Math & Question-answer pairs & \citet{mirzadeh2024gsmsymbolicunderstandinglimitationsmathematical}\\
    TheoremQA & 800 & Math &  Question-answer pairs & \citet{emnlp-2023-main}\\
    SVAMP & 1K & Math & Question-answer pairs & \citet{patel-etal-2021-nlp}\\
    MAWPS & 3.3K & Math & Collection of math word problems & \citet{koncel-kedziorski-etal-2016-mawps}\\
    AQUA-RAT & 100K & Math & Grade-school-math problems & \citet{ling-etal-2017-program}\\
    OmniMATH & 4428 & Math & Judged by OmniJudge, GPT-4o & \citet{gao2024omnimathuniversalolympiadlevel}\\
    Olympiad-Bench & 8476 & Math, Physics & Question-answer pairs & \citet{he2024olympiadbenchchallengingbenchmarkpromoting}\\
    Humaneval & Hundred & Code & Handwritten code & \citet{chen2021codex} \\
    APPS & 10K & Code & Question-code solution pairs & \citet{hendrycksapps2021}\\
    LiveCodeBench & 442 & Code & Question-code solution pairs & \citet{jain2024livecodebenchholisticcontaminationfree}\\ 
    MBPP & 1K & Code & Basic algorithmic and functional programming tasks & \citet{austin2021programsynthesislargelanguage}\\
    CommonsenseQA & 12K & Commonsense &  Multiple-choice question-answer pairs & \citet{talmor-etal-2019-commonsenseqa}\\
    StrategyQA & 2.8K & Commonsense & Question-answer pairs & \citet{geva-etal-2021-aristotle}\\
    ARC & 7.8K & Commonsense & Multiple-choice question-answering pairs & \citet{clark2018thinksolvedquestionanswering}\\
    5-hop ProntoQA & $\infty$ & Logic & Question-answer pairs  & \citet{saparov2023language}\\   
    AR-LSAT & 2,046 & Law & Question-answer pairs & \citet{zhong2021arlsatinvestigatinganalyticalreasoning}\\
    GPQA & Hundreds & Multiple domain & Question-answer pairs & \citet{rein2024gpqa} \\
    MMLU & 231K & Multiple domain & Reasoning-focused question-answer pairs & \citet{hendryckstest2021, hendrycks2021ethics}\\
    MMLU-pro & 12K & Multiple domain & Reasoning-focused question-answer pairs & \citet{wang2024mmluprorobustchallengingmultitask}\\
    \bottomrule
    \end{tabular}
\end{adjustbox}
\caption{Dataset on the recent reasoning methods}
\label{scc:dataset}
\end{table*}

\subsection{\textbf{ADAPT} Training Details}
\label{sec:appendix:adpatdetails}

\subsubsection{Custom Prompt Format}
\begin{prompt}
\texttt{"\{question\} Please provide the initial step towards resolving the question. This step may serve as a foundation but might not encompass the entire solution.\textbackslash n"}
\end{prompt}

\subsubsection{Training Parameter Setting}

\paragraph{Dataset}

The combined dataset is shuffled and split into $90\%$ for training and $10\%$ for testing, with the test portion further divided evenly into evaluation and held-out sets.

\paragraph{Tokenization}

Tokenization includes padding to a maximum length of 512 tokens and truncation when necessary.

\paragraph{Training}

Training is performed for $3$ epochs with a learning rate of $5\times10^{-6}$, per-device batch size of $4$, and gradient accumulation steps of $8$. We employ \texttt{bfloat16} precision, set \texttt{max\_grad\_norm} to $1.0$.

\end{document}